\documentclass[runningheads]{llncs}
\usepackage[colorlinks,citecolor=blue]{hyperref}
\usepackage{amsmath}
\usepackage{amssymb}
\usepackage{multirow}
\usepackage{tabularx}
\usepackage{rotating}
\usepackage{array}
\usepackage{graphicx}
\usepackage{authblk}
\usepackage{hyperref}

\begin{document}

\title{Feature Aggregation Network for Building Extraction from High-resolution Remote Sensing Images}
\titlerunning{FANet for Building Extraction from HR Remote Sensing Images}
\author{Xuan Zhou\inst{1}${\dagger}$, Xuefeng Wei\inst{1}${\dagger}$}

\authorrunning{X. Zhou et al.}

\institute{Institut Polytechnique de Paris, Rte de Saclay, 91120 Palaiseau, France 
\email{xuan.zhou@ip-paris.fr}\\
\email{xuefeng.wei@ip-paris.fr}}

\maketitle              
\renewcommand{\thefootnote}{}
\footnotetext[1]{$^{\dagger}$ Indicates equal contribution and Corresponding author.}

\begin{abstract}
The rapid advancement in high-resolution satellite remote sensing data acquisition, particularly those achieving sub-meter precision, has uncovered the potential for detailed extraction of surface architectural features. However, the diversity and complexity of surface distributions frequently lead to current methods focusing exclusively on localized information of surface features. This often results in significant intra-class variability in boundary recognition and between buildings. Therefore, the task of fine-grained extraction of surface features from high-resolution satellite imagery has emerged as a critical challenge in remote sensing image processing. In this work, we propose the Feature Aggregation Network (FANet), concentrating on extracting both global and local features, thereby enabling the refined extraction of landmark buildings from high-resolution satellite remote sensing imagery. The Pyramid Vision Transformer captures these global features, which are subsequently refined by the Feature Aggregation Module and merged into a cohesive representation by the Difference Elimination Module. In addition, to ensure a comprehensive feature map, we have incorporated the Receptive Field Block and Dual Attention Module, expanding the receptive field and intensifying attention across spatial and channel dimensions. Extensive experiments on multiple datasets have validated the outstanding capability of FANet in extracting features from high-resolution satellite images. This signifies a major breakthrough in the field of remote sensing image processing. We will release our code soon.
\keywords{Building extraction, Remote sensing image processing, Deep learning.}
\end{abstract}
\section{Introduction}
Modern remote sensing technology, with its sub-meter high-resolution satellite data, provides a deeper understanding of the surface of the earth. Especially the extraction of buildings, a core task of remote sensing image feature extraction, provides critical information for urban planning, population estimation and disaster assessment\cite{1}. However, as the resolution increases, the differences in shape, size, and style between buildings become more apparent. This intensifies intra-class differences and makes it difficult for the model to correctly distinguish between the land object background and the main body of the building. Moreover, factors such as trees and shadows can also reduce segmentation accuracy. Faced with the diversity and complexity of surface distribution, manual classification of land objects is time-consuming and expensive\cite{2}. Therefore, how to extract detailed surface features from high-resolution remote sensing images\cite{4,5} has become a major challenge in the field of remote sensing image processing.

Most of the existing work concentrates on building extraction methods from high-resolution remote sensing images using machine learning, and while substantial progress has been made, these methods are heavily reliant on manual feature design. With the evolution of deep learning, particularly the Fully Convolutional Networks (FCNs)\cite{6,7}, they have demonstrated remarkable progress in enhancing the accuracy and efficiency of building extraction from high-resolution remote sensing images through end-to-end network structures. However, when dealing with high-resolution, complex structures and patterns of remote sensing images, they still face challenges of insufficient global context information. Recently, many researchers have started to use methods based on Transformer, using its powerful global context information acquisition ability to effectively extract complex and diverse land object features. However, its fully connected self-attention mechanism can lead to the neglect of spatial structure information. This results in redundant attention in image processing tasks, leading to a decline in performance in fine spatial information tasks such as building boundary extraction.

We propose a novel Feature Aggregation Network (FANet), which uses a Pyramid Vision Transformer (PVT)\cite{8} in the encoder part of its new design. It effectively addresses attention map redundancy in conventional Transformer-based methods through multi-scale structural design. This revolutionary enhancement notably bolsters the accuracy and efficiency of our model in building extraction tasks. Specifically, the Aggregation Module strengthens the local information in the global features already extracted by the Transformer through spatial and channel information filtering. More precisely, its focus lies in optimizing and supplementing the Transformer's global features from a local perspective. The Difference Elimination Module enhances image comprehension by fusing features at different levels, thereby facilitating interpretation from both global and local perspectives. This effectively compensates for the Transformer's limitations in understanding spatial relationships. Concurrently, the Receptive Field Block and Dual Attention Module augment the model's perception of global and local features by expanding the receptive field and intensifying attention across both spatial and channel dimensions. The Fusion Decoder is responsible for effectively fusing features at high and low layers to output detailed land object extraction results. Extensive experiments on several datasets demonstrate the efficiency of our proposed FANet model in extracting detailed features from high-resolution remote sensing images. The multi-module cooperative design of the model marks a substantial advance in the field of remote sensing image processing. The main contributions of this research are:
\begin{enumerate}
\item We propose a FANet framework that enhances the accuracy of landmark building segmentation in high-resolution satellite remote sensing images.
\item The feature aggregation and dual attention modules, designed to filter information and enhance spatial and channel data, boost the model's accuracy and efficiency in extracting buildings from remote sensing images.
\item Experimental results show that FANet outperforms most state-of-the-art models on challenging datasets, demonstrating the effectiveness and robustness of our method in handling complex remote sensing image data.
\end{enumerate}

\section{Related Work}
Building extraction has made significant strides in research and has played an important role in various fields, such as human activities and socio-economics among others. Earlier studies primarily relied on manually designed features, such as shape,  context and shadow indices to identify buildings \cite{9,11,12}. Subsequent research \cite{13} began to introduce endmembers and associated filters to separate buildings from the background. With the advent of deep learning technologies, techniques such as Fully Convolutional Neural Networks (CNNs) \cite{14} have been introduced into building extraction, significantly improving the execution of this task \cite{15}. For instance, methods such as deep neural networks based on autoencoders \cite{16} and CNNs based on a single path \cite{17} have shown excellent performance in this regard. Despite these methods have proven the effectiveness of using deep neural networks for building extraction, they tend to overlook the impact of building layout changes.

To solve the aforementioned problem, researchers have proposed several methods for multi-scale feature extraction, such as parallel networks \cite{19}, and pyramid-based methods \cite{18}. These methods can extract building features from different perspectives and scales, but may overlook the interaction between information at different scales. Meanwhile, to enhance the feature recognition capabilities of CNNs, some research began to introduce attention mechanisms \cite{20,21}, while others have achieved higher classification accuracy by integrating the advantages of different network structures or modules \cite{22,23}. However, due to a lack of attention to edge details, the building contours extracted by these methods are often irregular. Recent research has begun to seek methods that reduce the amount of annotation work without sacrificing accuracy. For example, some of the latest methods \cite{24} attempt to model the features of building edges and interiors more accurately through a coarse-to-fine hierarchical training strategy. These methods have been able to model the overall layout and detailed information of buildings effectively, capturing the detailed information inside buildings while ensuring edge accuracy.

However, we note that when dealing with high-resolution satellite remote sensing images, the accuracy of the aforementioned methods in dealing with the edge details of buildings is not ideal, often extracting irregular building contours. To address this, we propose a Feature Aggregation Network (FANet) with a Pyramid Vision Transformer (PVT) \cite{8} as the encoder. PVT is a network that employs vision transformers, capable of effectively combining global and local information to improve feature extraction and classification performance. With PVT, our FANet can effectively capture the detailed information inside buildings while ensuring edge accuracy, providing an effective solution for high-precision building extraction.

\begin{figure}[t!]
\includegraphics[width=\textwidth]{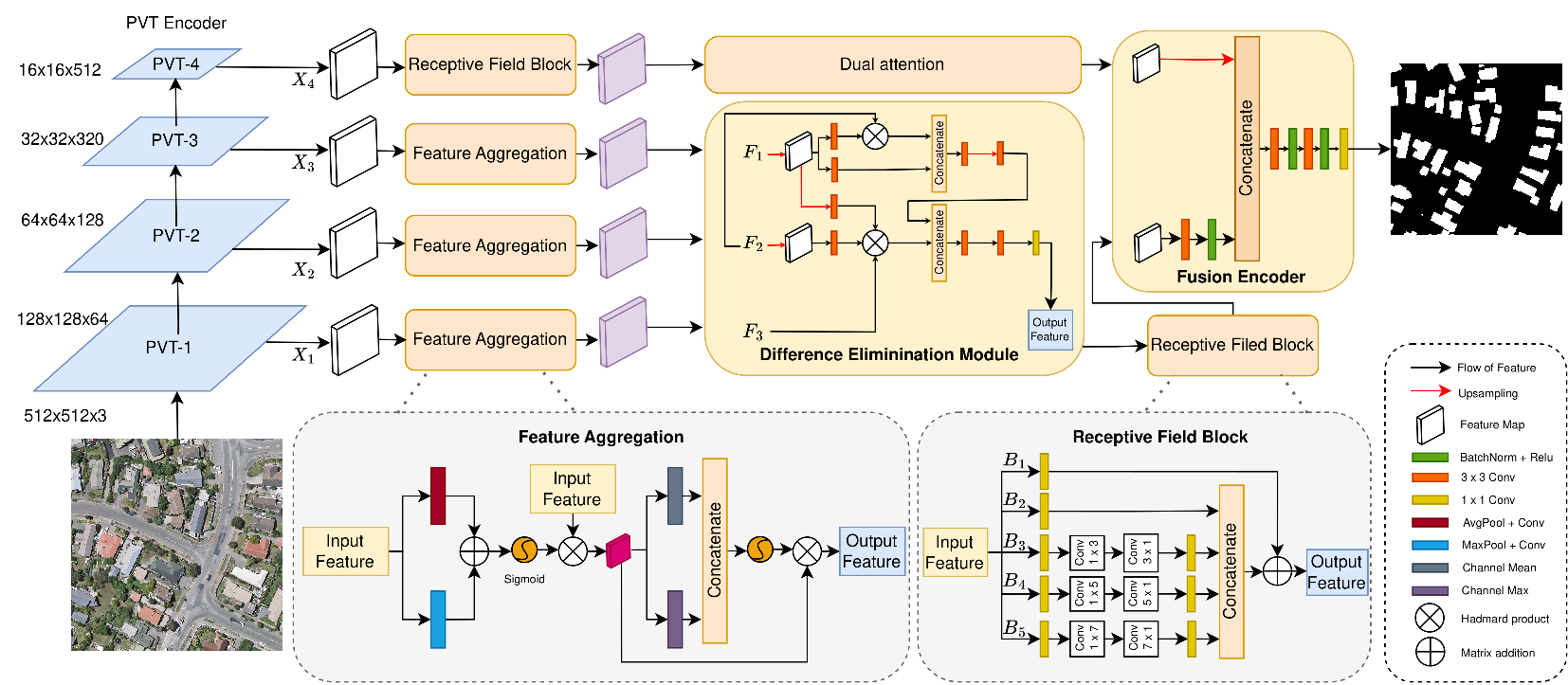}
\caption{The proposed Feature Aggregation Network (FANet) workflow. Starting with the Pyramid Vision Transformer for multi-scale feature extraction, the process seamlessly progresses through feature enrichment, integration, receptive field expansion, and dimension amplification, concluding with the Fusion Decoder outputting the final building segmentation.} \label{fig1}
\end{figure}

\section{Methodology}\label{First}
We propose the Feature Aggregation Network (FANet), an innovative approach to the fine-grained extraction of buildings from high-definition remote sensing imagery. As illustrated in Fig. \ref{fig1}, FANet is designed around the concept of feature enrichment and integration. It starts with the extraction of multi-scale, long-range dependencies using the Pyramid Vision Transformer encoder. To these initial features, the Feature Aggregation Module provides further enrichment and the Difference Elimination Module integrates low-level details. By expanding the receptive field through the Receptive Field Block and intensifying the spatial and channel dimensions with the Dual Attention Module, FANet achieves a holistic understanding of the image data. The final building segmentation is realized by the Fusion Decoder. The subsequent sections provide a comprehensive exploration of these components and their intricate interplay.

\subsection{Transformer Encoder}
Given the extensive spatial coverage of remote sensing images, a Pyramid Vision Transformer (PVT) is employed to effectively extract features. The transformer-based backbone\cite{25} processes an input image $I \in \mathbb{R}^{H \times W \times 3}$, generating pyramid features $X_i \in \mathbb{R}^{\frac{H}{2^{i+1}} \times \frac{W}{2^{i+1}} \times C_i}$, where $C_i \in {64, 128, 320, 512}$ and $i \in {1, 2, 3, 4}$. The channels of the low-layer features $X_1$, $X_2$, and $X_3$ are then acquired via convolutional units and passed to the Feature Aggregation Module (FAM).

This PVT model employs patches of different scales which are fed into Transformers at multiple levels, thereby forming a pyramid-like structure. This approach aids in capturing multi-scale image features, enhancing the model's performance while reducing the computational burden associated with high-resolution images. Specifically, we adopt PVTv2\cite{36}, an improved version of PVT, which possesses a stronger feature extraction capability. Adapting PVTv2 to the task of image segmentation, we discard the last classification layer and design a segmentation head based on the multi-scale feature maps $X_1$, $X_2$, $X_3$, and $X_4$. These feature maps provide both lower-layer appearance information of building images and high-layer features.

\subsection{Feature Aggregation Module}
The Feature Aggregation Module (FAM) is employed to refine and enrich the low-layer features ($X_{1}$, $X_{2}$, $X_{3}$) in both spatial and channel dimensions. By applying Global Adaptive Average Pooling (GAvgPool) and Global Adaptive Max Pooling (GMaxPool), we generate channel response maps. The aggregation of these maps, after passing through a Sigmoid function, generates a probability map, $R'$. Subsequently, channel-level fusion is achieved by the element-wise multiplication of the input $F$ and $R'$, resulting in the feature map $F_{\text{channel}}$. The process can be mathematically expressed as:
\begin{equation}
F_{\text{channel}} = \text{Sigmoid}(\text{GAvgPool}(F) + \text{GMaxPool}(F)) \cdot F,
\end{equation}
where $F_{\text{channel}} \in \mathbb{R}^{B \times C \times H \times W}$. This method ensures a balanced representation of global and salient features, leading to a comprehensive channel-wise feature profile.

Next, to perform spatial-level feature aggregation, we use the derived feature map $F_{\text{channel}}$. We compute mean and max responses across all channels, concatenate them, and pass through a convolution and a Sigmoid function. The resulting probability map, $T'$, is used for spatial-level fusion, yielding $F_{\text{spatial}}$ as follows:
\begin{equation}
F_{\text{spatial}} = \text{Sigmoid}(\text{concat}(\text{Mean}(F_{\text{channel}}), \text{Max}(F_{\text{channel}}))) \cdot F_{\text{channel}},
\end{equation}
where $F_{\text{spatial}} \in \mathbb{R}^{B \times C \times H \times W}$. This procedure enables the model to capture average representation and distinctive spatial characteristics, thus providing a more enriched spatial feature description.

\subsection{Feature Refinement via Difference Elimination Module and Receptive Field Block}
In our method, the Difference Elimination Module (DEM) and the Receptive Field Block (RFB) play critical roles in refining the initial low-level features and crafting a diverse, unified feature representation as shown in Fig.\ref{fig1}.
Upon processing by the Feature Aggregation Module (FAM), initial low-level features are transformed into enhanced features ($F_1$, $F_2$, $F_3$). To reconcile differences between adjacent features, these undergo upsampling, convolution, and element-wise multiplication with higher layer maps, resulting in a composite feature representation.
These maps are then funneled through the RFB, comprising five branches with adaptable kernel sizes and dilation rates in branches $k>2$, capturing multi-scale information. The outputs from the last four branches are concatenated and element-wise added to the first branch's output, fostering rich feature interactions.
The DEM and RFB synergistically address layer discrepancy issues and enhance the overall feature representation's diversity. The ensuing section will delve into the application of the Dual Attention Module that further refines these composite features, leading to more effective model performance.

\subsection{Dual Attention Module for Enhanced Feature Interactions}
The Dual Attention Module (DAM) refines high-layer features by capturing interactions across spatial and channel dimensions. It is designed to harness the dependencies inherent in the features along these dimensions, thus augmenting feature representation and enhancing the model's overall interpretive capacity.

The DAM takes an input feature map $A \in \mathbb{R}^{B \times C \times H \times W}$ and generates two new feature maps, $B$ and $C$. These are utilized to compute a spatial attention map $S \in \mathbb{R}^{N \times N}$. Concurrently, a separate convolution operation on $A$ interacts with $S$ to produce the spatially refined feature map $E$. This is mathematically encapsulated as follows:

\begin{equation}
E_{j} = \gamma \sum_{i=1}^{N} \left( \frac{exp(B_{i} \cdot C_{j})}{\sum_{i=1}^{N} exp(B_{i} \cdot C_{j})} D_{i} \right) + A_{j}, \tag{3}
\end{equation}
where $\gamma$ is a learnable weight parameter that controls the trade-off between the original and the spatially-attended features, initialized as 0. This allows the model to progressively learn the optimal balance as training progresses.

In parallel, the module exploits channel-wise interdependencies by generating a channel attention map $X \in \mathbb{R}^{C \times C}$ directly from $A$. This map interacts with $A$ to yield the channel-refined feature map $M$:

\begin{equation}
M_{j} = \beta \sum_{i=1}^{C} \left( \frac{exp(A_{i} \cdot A_{j}^{T})}{\sum_{i=1}^{C} exp(A_{i} \cdot A_{j}^{T})} A_{i} \right) + A_{j}, \tag{4}
\end{equation}
where $\beta$ is another learnable weight parameter, also initialized as 0. Similar to $\gamma$, it controls the mix between the original and the channel-attended features, letting the model learn the optimal balance during training.

The module concludes by combining the spatially and channel-wise refined feature maps, $E$ and $M$, using an element-wise addition. A 1x1 convolution operation follows, reducing dimensions to complete the fusion enhancement of features. Consequently, the DAM effectively enhances the model's overall capability to understand complex scenes by promoting richer feature interdependencies.

\subsection{Fusion Decoder and Loss Function}
Our Fusion Decoder, shown in Fig.\ref{fig1}, integrates high-level global contexts with detailed local features, optimizing segmentation. High-level features, providing holistic target understanding, are resized to match low-level features, ensuring a comprehensive fusion. After merging, these features undergo convolutional refinement. A 1x1 convolution yields a predicted segmentation map, resized to the input image dimensions for the final result. The decoder maintains high-level contexts and leverages low-level details, enhancing segmentation. We use the Binary Cross Entropy (BCE) loss to measure consistency between predictions and ground truth.

\section{Experiments}
\begin{figure}[t!]
\includegraphics[width=\textwidth]{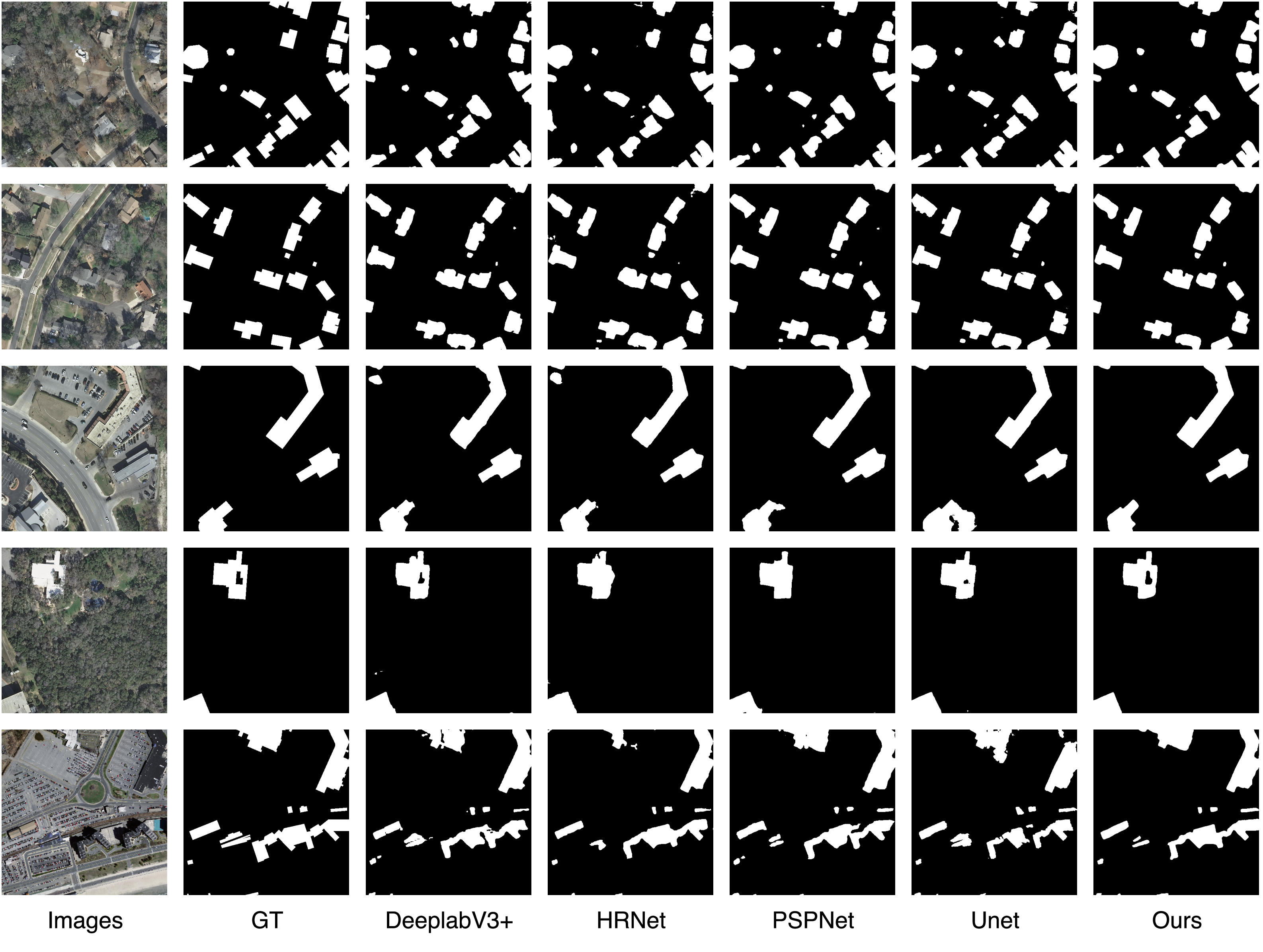}
\caption{Visual comparison between our results and those of state-of-the-art methods. The first and second columns represent the building images and the corresponding ground truth, respectively. Columns 3 to 7 display the results generated by UNet \cite{26}, PSPNet \cite{27}, Deeplabv3+ \cite{28}, HRNet \cite{29}, and Ours, respectively.} \label{fig7}
\end{figure}

\subsection{Datasets}
To assess the efficiency of our method, we experimented with three public datasets, namely, the WHU Building dataset \cite{33}, the Massachusetts Building dataset \cite{34}, and the Inria Aerial Building Dataset \cite{35}. The WHU Building dataset contains approximately 220,000 buildings from aerial images. We partitioned the 8189 images of 512 x 512 pixels into a training set (4736 images), a validation set (1036 images), and a test set (2416 images). The Massachusetts Building dataset, which includes 151 aerial images from the Boston area, was divided post non-overlapping cropping into 512 x 512 pixels, resulting in 3076 training images, 100 validation images, and 250 test images. Finally, the Inria Aerial Image Labeling Dataset, covering an area of 810 km\(^2\) across five cities, was split into 8271 training images and 1600 test images after removing training images without buildings. These three datasets, with their diverse geographical contexts, provide a rigorous testing environment for our proposed method.

\subsection{Implementation Details}
In our model, input image data and label data were cropped to a size of 512 x 512 pixels. The cropped images underwent data augmentation techniques, including random horizontal flipping and random Gaussian blurring, to generate the preprocessed dataset. Subsequently, all models were trained using the same parameter settings and environment. Our model was developed under the framework of Pytorch 1.8.1 and cuda 11.1, with the hardware of a single GeForce RTX 3090 with 24GB of computation memory. The initial learning rate was set to $1 \times 10^{-4}$, and the Adam optimizer was employed. The learning rate was decayed by a factor of 10 every 50 epochs, and the training process lasted for a total of 100 epochs. To quantitatively evaluate the performance of the proposed method, this study employs four metrics, namely Precision, Recall, F1-score, and Intersection over Union (IoU).
\subsection{Comparison with Other State-of-the-Art Methods}\label{11:seventh}
To evaluate the effectiveness of the proposed method, we compared our approach with other state-of-the-art methods, including UNet \cite{26}, PSPNet \cite{27}, Deeplabv3+ \cite{28}, HRNet \cite{29}, BOMSNet \cite{30}, LCS \cite{31}, and MSNet \cite{32}. These comparisons were conducted when applied to the WHU Building dataset, the Massachusetts Building dataset, and the Inria Aerial Building Dataset. We performed a visual qualitative evaluation of the experimental results, as shown in Fig.\ref{fig7}. As can be observed from the results, compared to other state-of-the-art methods, our approach yielded superior results in building extraction.  
Quantitative evaluation results are shown in Table \ref{tab:table}.
\begin{table}[htp]
\centering
\caption{Experimental Results for Various Datasets, The bolded data shown in the table indicates the best data on the corresponding metric and the data with underline indicates the second best one on the corresponding metric. }
\label{tab:table}
\begin{tabularx}{\textwidth}{>{\centering\arraybackslash}p{3cm} >{\centering\arraybackslash}p{3cm} X X X X}
\hline
 Dataset &  Methods &  IoU &  F1 &  Pre &  Recall \\
\hline
\multirow{6}{=}{ Massachusetts Building Dataset} &  UNet &  67.61 &  80.68 &  79.13 &  \underline{82.29} \\
&  PSPNet & 66.52 & 79.87 & 78.53 & 81.26 \\
& Deeplabv3+ &  69.23 &  81.82 &  84.73 & 79.10 \\
&  HRNet &  69.58 &  \underline{82.01} &  \underline{85.06} &  79.17 \\
&  MSNet &  \underline{70.21} &  79.33 &  78.54 &  80.14 \\
&  \textbf{Ours} &  \textbf{73.35} &  \textbf{84.63} &  \textbf{86.45} &  \textbf{82.87} \\
\hline
\multirow{7}{=}{ Inria Aerial Building Dataset} &  UNet &  74.40 &  85.83 &  86.39 &  84.28 \\
&  PSPNet &  76.8 &  86.88 &  87.35 &  86.4 \\
&  Deeplabv3+ &  78.18 &  87.75 &  87.93 &  87.58 \\
&  HRNet &  \underline{79.67} &  \underline{88.68} &  \underline{89.82} &  \underline{87.58} \\
&  BOMSNet &  78.18 &  87.75 &  87.93 &  87.58 \\
&  LCS &  78.82 &  88.15 &  89.58 &  86.77 \\
&  \textbf{Ours} &  \textbf{81.05} &  \textbf{89.53} &  \textbf{90.49} &  \textbf{88.60} \\
\hline
\multirow{6}{=}{ WHU Building Dataset} &  UNet &  85.51 &  92.19 &  91.86 &  92.52 \\
&  PSPNet &  86.68 &  92.55 &  92.25 &  92.86 \\
&  Deeplabv3+ &  85.78 &  92.35 &  93.45 &  91.27 \\
&  HRNet &  87.85 &  93.21 &  94.22 &  92.23 \\
&  MSNet &  \underline{89.07} &  \underline{93.96} &  \textbf{94.83} &  \underline{93.12} \\
&  \textbf{Ours} &  \textbf{90.01} &  \textbf{94.74} &  \underline{94.50} &  \textbf{94.98} \\
\hline
\end{tabularx}
\end{table}

The rows represent different tested methods, and the columns represent different evaluation metrics. It can be observed that the method we proposed achieved the best performance across all three datasets. 
On the Massachusetts Building dataset, our proposed method surpassed the second-best method (i.e., MSNet) by approximately 4.47\% in IoU, and surpassed the second-best method (i.e., HPNet) by approximately 3.19\% and 1.63\% in terms of F1-score and Precision, respectively. On the Inria Aerial Building Dataset, compared to HPNet, the IoU and Recall metrics of our proposed method increased by approximately 1.73\% and 1.16\%, respectively. On the WHU Building dataset, compared to MSNet, the Recall of our proposed method improved by approximately 2\%. Experimental comparison with different methods indicates that our proposed method outperformed others on all three datasets.
\subsection{Ablation study}
In our investigation, an exhaustive ablation study validates the efficacy of pivotal components within the FANet architecture, namely, the Feature Aggregation Module (FAM), Receptive Field Block (RFB), Dual Attention Module (DAM), and Difference Elimination Module (DEM). These experiments were carried out on the Massachusetts Building Dataset, employing standard benchmarks such as IoU, F1-score, Precision, and Recall.
\begin{figure}[t!]
\begin{center}
\includegraphics[width=0.7\textwidth]{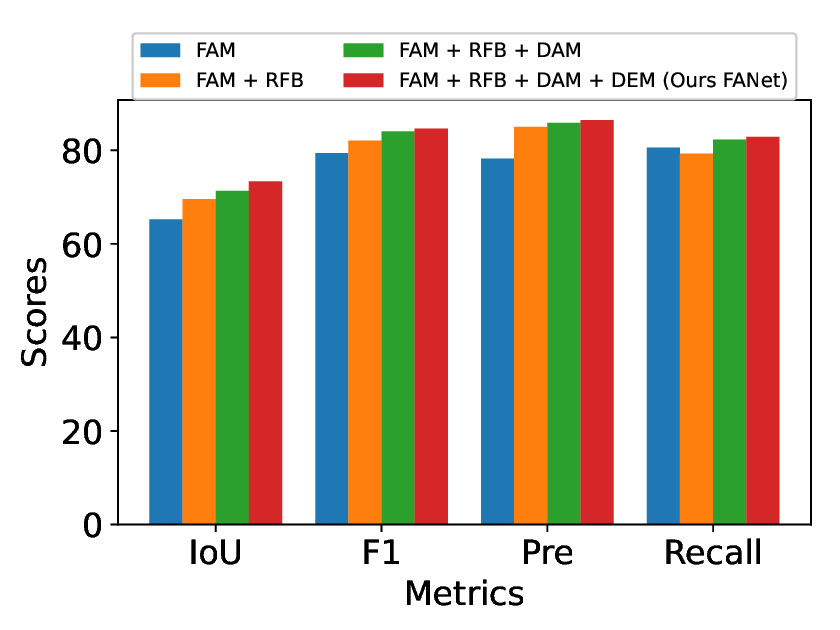}
\end{center}
\caption{The results of ablation experiment on Massachusetts Building Dataset.} \label{fig4}
\end{figure}

As demonstrated in Fig.\ref{fig4}, FANet excels across all evaluation metrics. Preliminary experiments with solely FAM displayed a notable enhancement over the baseline model, thereby attesting FAM's effectiveness in information filtering. Upon integrating the RFB with FAM, the model's performance underwent further improvement, underlining the RFB's indispensable role in receptive field expansion. Models lacking the DEM performed sub-optimally in comparison to FANet, indicating that feature fusion amplifies the model's performance. Moreover, FANet surpassed the FAM+RFB+DAM configuration by approximately 2.8\% in terms of IoU, accentuating the importance of feature fusion. In summary, the ablation study effectively highlighted the potency of each module within FANet. The stepwise integration of each module led to significant performance enhancements, underscoring their integral role within the network.

\section{Conclusion}
In this work, we propose a novel Feature Aggregation Network (FANet) for the fine-grained extraction of buildings in high-resolution satellite remote sensing images. This is to address the challenges posed by the variance in shape, size, and style among buildings, and the complex ground conditions that make it difficult to distinguish the main body of buildings. Experimental results on three open-source datasets have validated the effectiveness of the proposed network. FANet can accurately extract the boundaries of buildings in complex environments such as shadows and tree occlusions and achieve the complete extraction of buildings of different scales. For future research in building extraction, designing a robust model trained with less data presents a promising direction.

%
%
%

\end{document}